\documentclass[conference]{IEEEtran}
\IEEEoverridecommandlockouts
\usepackage{cite}
\usepackage{amsmath,amssymb,amsfonts}
\usepackage{algorithmic}
\usepackage{graphicx}
\usepackage{textcomp}
\usepackage{xcolor}
\usepackage{subfigure}
\usepackage{multicol}
\usepackage{multirow}
\usepackage{array} 
\usepackage{booktabs} 
\def\BibTeX{{\rm B\kern-.05em{\sc i\kern-.025em b}\kern-.08em
    T\kern-.1667em\lower.7ex\hbox{E}\kern-.125emX}}
\begin{document}

\title{Federated Learning in Big Model Era: Domain-Specific Multimodal Large Models\\

\thanks{Identify applicable funding agency here. If none, delete this.}
}
\author{Zengxiang Li$^{1}$,  Zhaoxiang Hou$^{1}$, Hui Liu$^{1}$, Ying Wang$^{1}$,Tongzhi Li$^{1}$,  Longfei Xie$^{1}$, Chao Shi$^{1}$,Chengyi Yang$^{1}$,  \\
Weishan Zhang$^{2}$, Zelei Liu$^{3}$, Liang Xu$^{4}$ \\
\normalsize $^{1}$ENN Group Co.,Ltd., China\\
\normalsize $^{2}$China University of Petroleum, China\\
\normalsize $^{3}$Unicom (Shanghai) Industrial Internet Co.,Ltd., China\\
\normalsize $^{4}$Qingdao Windaka Technology Co.,Ltd., China\\
\normalsize e-mail:\{lizengxiang,houzhaoxiang,liuhuiau,wangyingbj,litongzhi,
xielongfei,shichaog,yangchengyia\}@enn.cn, \\zhw@s.upc.edu.cn, liuzl231@chinaunicom.cn, xuliang.upc.edu@gmail.com
}

	
\maketitle

\begin{abstract}
With the tremendous success of large language models represented by ChatGPT, artificial intelligence has ushered in a new wave and entered the era of big models. Multimodal data, which can comprehensively perceive and recognize the physical world, has become an essential path towards general artificial intelligence. However, multimodal large models trained on public datasets often underperform in specific industrial domains. This paper proposes a multimodal federated learning framework that enables multiple enterprises to utilize private domain data to collaboratively train large models for vertical domains, achieving intelligent services across scenarios. The authors discuss in-depth the strategic transformation of federated learning in terms of intelligence foundation and objectives in the era of big model, as well as the new challenges faced in heterogeneous data, model aggregation, performance and cost trade-off, data privacy, and incentive mechanism. The paper elaborates a case study of leading enterprises contributing multimodal data and expert knowledge to city safety operation management , including distributed deployment and efficient coordination of the federated learning platform, technical innovations on data quality improvement based on large model capabilities and efficient joint fine-tuning approaches. Preliminary experiments show that enterprises can enhance and accumulate intelligent capabilities through multimodal model federated learning, thereby jointly creating an smart city model that provides high-quality intelligent services covering energy infrastructure safety, residential community security, and urban operation management. The established federated learning cooperation ecosystem is expected to further aggregate industry, academia, and research resources, realize large models in multiple vertical domains, and promote the large-scale industrial application of artificial intelligence and cutting-edge research on multimodal federated learning.
\end{abstract}

\begin{IEEEkeywords}
multimodal model, federated learning, big model
\end{IEEEkeywords}

\section{Introduction}
Thanks to several decades of rapid development in data, computing power, and algorithms, artificial intelligence has entered a new era of enthusiasm. Large-scale language models such as ChatGPT\cite{ouyang2022training} have achieved incredible effects, particularly in understanding human intentions, context learning, and zero-shot learning for general tasks. The achievements have swiftly influenced the field of computer vision as well. Works like SAM\cite{kirillov2023segment} and UniDetector\cite{wang2023detecting} have demonstrated capabilities in open-world object recognition and segmentation. This surge in research breakthroughs and astonishing intelligent abilities has
instilled the hope of achieving universal artificial intelligence.

To realize general artificial intelligence, intelligent agents need to be capable of processing and comprehending multimodal information while interacting with the real world~\cite{DBLP:journals/corr/abs-2004-10151}. Multimodality has an innate advantage in information fusion to cover everything from traditional sensory recognition to complex cognitive reasoning and decision making tasks. For instance, by adding lip motion image modal data, some vague pronunciations can be accurately recognized, in the autonomous driving domain, by comprehensively utilizing sensor and camera signals, the perception of road conditions is enhanced, in the field of robot control, RGB-D and 3D point clouds, due to the introduction of depth information, have given intelligent entities a concept of "distance" in their perception of the environment.

Multimodal data, which includes text, image, video, 3D point clouds, audio, vibration signals, and more, inherently possess characteristics such as heterogeneity, interactivity, and correlations. Effectively combining diverse sources of multimodal data using cutting-edge large-scale model techniques and deeply exploring the relationships among various modalities to integrate information for intelligent emergent effects has rapidly become one of the most prominent research topics. The mainstream approaches includes fusing multimodal feature through cross-attentions such as CLIP\cite{radford2021learning},ALBEF\cite{li2021align}, and BLIP\cite{li2022blip} and incorporating multimodal information into a powerful large language model such as OFA\cite{wang2022ofa}, Flamingo\cite{alayrac2022flamingo}, Palm-E\cite{driess2023palm} .

\begin{figure*}[t]
\centering
\includegraphics[width=0.8\textwidth]{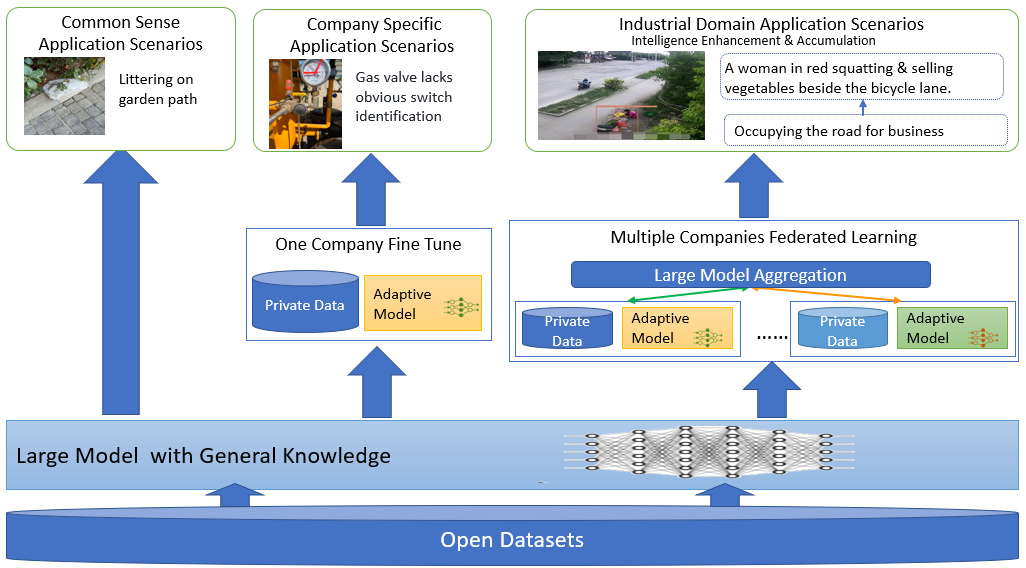}
\caption{Private data fine-tuning and federated learning amongst multiple companies are essential to transform a general large model to domain-specific model for real-world industrial application adoption.}
\label{fig:FLLLM}
\end{figure*}

Most of the multimodal large models provided by AI giants and start-ups are based on public datasets. They could achieve high performance on the validation datasets in experimental environments and could adapt to some application scenarios requiring common sense only. However, these models often fall short of expectations in specific industrial applications. To enhance their performance, fine-tuning of these general large models with enterprise private data, alongside the integration of domain expertise, becomes essential to a successful adoption. This process enables the creation of domain-specific large models, enriching domain understanding and producing more specialized and accurate outputs. The method of fine-tuning based on large models can integrate domain knowledge on top of world knowledge. Compared with the traditional small models trained end-to-end in special domains, it has better generalization capability and can quickly adapt to open industrial application environments. This is conducive to the adoption of industrial intelligence in large amounts of application scenarios with lower cost.

As shown in Fig \ref{fig:FLLLM}, by fine-tuning the general large model based on its proprietary data, a company can enhance model intelligent capabilities with integrated domain-knowledge, such as special device recognition and safety inspection rules, in its existing application scenarios.  However, the application scenarios, proprietary data and expert knowledge of an individual company is often insufficient to cover the entire industry domain. Only through ecological cooperation, a number of companies could contribute complementary and multimodal data, as well as different aspects of expert knowledge, and thus jointly train a domain-specific large model covering as many as scenarios in the industry domain. This model can support various intelligent tasks complementing and stacking each other, and have stable and reliable performance in various practical application environments. For instance, models jointly trained by energy companies, household appliance companies, and healthcare companies can provide household customers with intelligent services for low-carbon and healthy living. 

\begin{figure*}[t]
\centering
\includegraphics[width=0.8\textwidth]{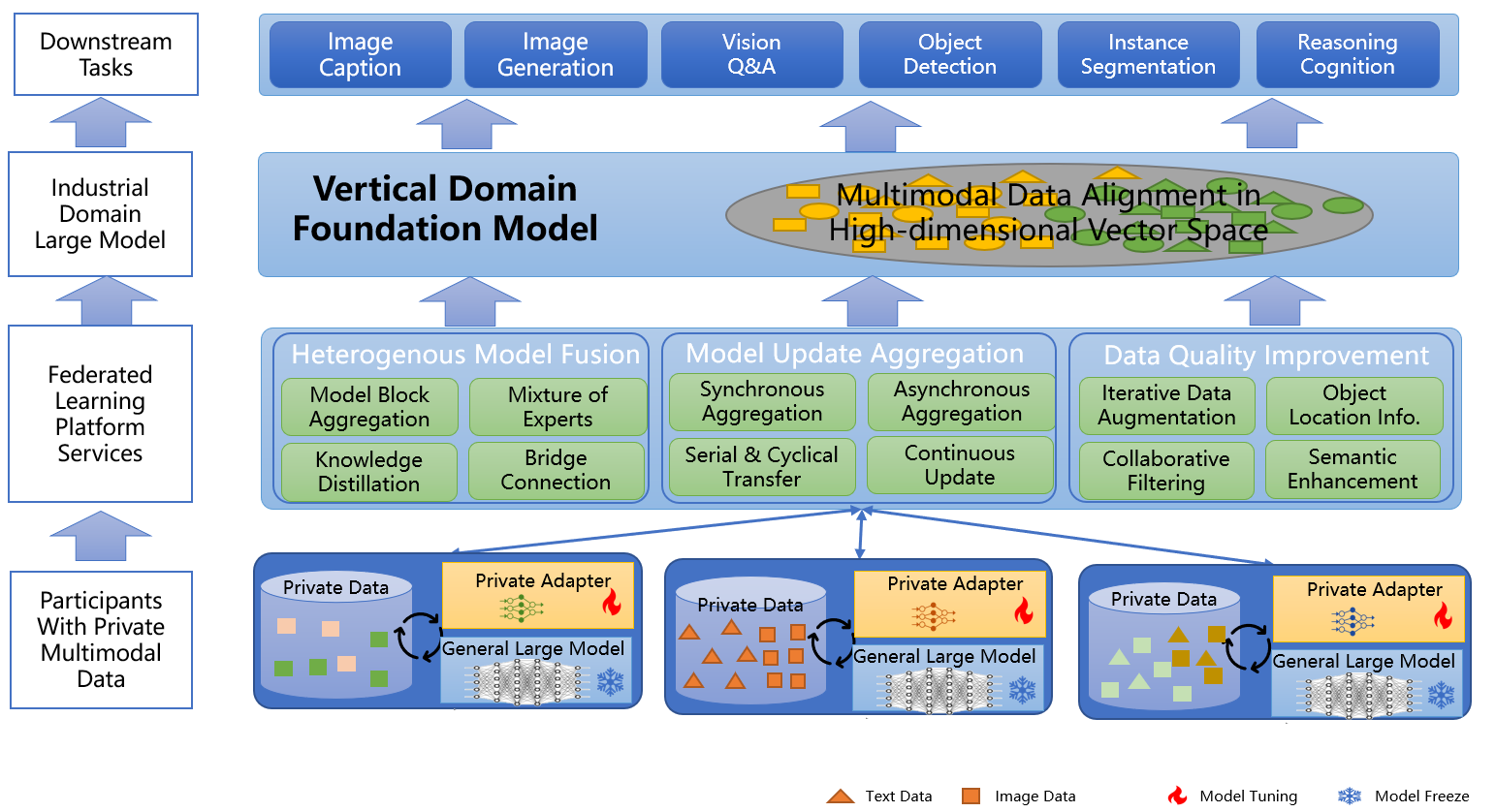}
\caption{Federated learning framework in big model era: enables companies collaboratively train domain-specific models by harnessing the capabilities of an open general large model while utilizing their private multimodal data.}
\label{fig:Framework}
\end{figure*}

Due to data privacy and asset value concerns, companies are reluctant to share data with each other. Federated learning(FL)\cite{mcmahan2017communication} provides a new AI paradigm to enable companies to jointly train a model while keeping data within their local computing resources. Traditional federated learning leverages similar or relevant data from multiple companies to train a global model from scratch, for the purpose of improving higher accuracy within the specific application scenarios. In big model era, the concept and definition of federated learning should be extended and reconsidered to enable aforementioned ecological cooperation aiming at domain specific large models for industrial applications. It starts by leveraging the power of pre-trained large models using vast public datasets as the intelligence foundation, and jointly fine-tuning the foundation model using multimodal data from various companies. The multimodal large model generated by federated learning not only has the intelligence capabilities of each participant, but also integrates and fuses their intelligence. As a result, it enables all-encompassing intelligent capabilities across industrial scenarios and expanding the intelligent application boundaries for individual participants. As shown in Figure~\ref{fig:FLLLM}, thanks to the collaborative intelligence among a city operation management company, a residential security company and an energy company, the multimodal model could provide informative description in terms of the environment background (bicycle lane along a arterial road), nearby energy infrastructure (e.g., natural gas pipeline and well cover), as well as people's feature and behaviour (e.g, a woman in read squatting and selling vegetables).

In this paper, we propose a foundational technical framework for multimodal federated learning that enables multiple enterprises to collaboratively train large-scale models in vertical domains using proprietary data sources. The paper is organized as follows, Section~\ref{FrameworkSection} illustrates the details of federated learning multimodal model framework and corresponding research directions to explore. Section~\ref{CaseStudySection} presents a smart city vision-language model case study, including federated learning framework deployment and data enhancement methods. Section~\ref{ConclusionSection} concludes the paper.


\section{Federated Learning Framework for Domain-Specific Multimodal Large Models}
\label{FrameworkSection}

\begin{table*}[htbp] 
  \centering
  \caption{Comparison of Federated Learning in Small Model Era and Large Model Era}
  \begin{tabular}{p{0.10\linewidth}p{0.4\linewidth}p{0.4\linewidth}}
    \toprule
    Dimension & Federated Learning in Small Model Era & Federated Learning in Large Model Era \\
    \midrule
    Intelligence Foundation & Training models from scratch or randomly initiated parameters. & Fine-tuning based on pre-trained general large models. \\
    Intelligent \newline   Objectives & Multiple companies collaboratively training global models to enhance accuracy in specific scenarios within closed environment. & Multiple companies collaboratively train domain-specific large models, supporting comprehensive intelligent services for industry applications and open environment. \\
    Federated Data & Single modality, spatiotemporal alignment, similarity/correlation data with statistical heterogeneity. & Diverse modalities and heterogeneous forms, challenging sample and spatiotemporal alignment, significant data quality and integrity variations. \\
    Federated \newline Models & High-frequency synchronized aggregation and iterative refinement. Data-similar clustering, model-adaptive aggregation algorithms to address data heterogeneity and improve aggregation model performance. & Participants contribute advantageous modules and small portions of parameters in large model structure, through asynchronous aggregation, iterative updates, or continuous learning, fine-tuning general large models collaboratively. \\
    Performance Optimization & Using techniques such as model quantization compression, adaptive aggregation module selection, and knowledge distillation to reduce communication overhead. & Strategies for large model collaborative fine-tuning need to adapt to ecosystem partners' variations in data and computational resources, balancing model performance and cost overhead. \\
    Privacy \newline Protection & Employing techniques like multi-party secure computation, homomorphic encryption, differential privacy, and trusted execution environments to prevent leakage of sensitive information during model transmission and aggregation. & Privacy leakage risks exist in both model transmission and large model output. Requires careful and comprehensive balance between security levels and additional costs. \\
    Incentive Mechanism & Evaluating data quality and contribution to the model by each party as the foundation for fair distribution and stimulating ecosystem development. & Evaluating specific characteristics, real-time nature, and professionalism of participant data, as well as their collaborative contributions to improving large model performance in vertical domains. \\
    \bottomrule
  \end{tabular}
  \label{tab:comparison}
\end{table*}

Table~\ref{tab:comparison} makes the comparison between traditional federated learning(i.e., in small model era) and that in big model era. Traditional federated learning jointly train a global model from scratch for the purpose of enhance model accuracy within a predefined application scenarios. Nowadays, federated learning aims at a domain-specific model based on the foundational intelligence of a general large model. Since the application target and generation method of intelligent product have been fundamentally changed in big model era, federated learning faces new challenges to tackled.
\begin{itemize}
\item Data Multi-modality and Heterogeneity: Companies own data in different modalities, exhibiting diverse data format and statistical distributions, considerable differences in data quality and completeness and the data samples and features could not be aligned explicitly.  
\item Model Structure and Size: The amount of model parameters grows exponentially, involving the various deep learning networks and complex structures. Traditional federated learning aggregates updates of the entire model which leads to substantial computational and network communication costs. However, a company with limited local data usually affects a small portion of the well-structured general large model. Hence, the aggregation strategy should decide ``what to aggregate" and ``how to aggregate" taking the trade-off between model training cost and performance into account.

\item Data Privacy and Security: According to large model aggregation strategy, the potential privacy leakage risks should be re-assessed and cost-effective privacy preserving computation should be adopted. In addition, measurements should be taken to prevent the potential exposure of company private data through generative outputs of large models.

\item Incentive Mechanisms and Collaboration: Due to increasing model size and complexity, it becomes even more challenging to evaluate data value and model contributions of participants. Data uniqueness and domain knowledge become more important, which bring added values transforming general large model to domain-specific model.     

\end{itemize}

In this paper, we propose a novel federated learning framework, as illustrated in Fig \ref{fig:Framework}, which enables multiple companies to collaboratively train domain-specific large models using their proprietary data sources. Federated learning participants have private data of different modalities. The client can use various methods to implement heterogeneous models, such as using the LoRA method to train a small amount of network parameters for the model, training only a specific expert module (MOE), transferring feature maps through knowledge distillation, or training bridge networks for different modal encoders.The server fuses the heterogeneous models from other participants using a flexible model aggregation method and distributes the fused model for iterative training. With the help of the fused model’s capability, the data quality of the participants is continuously improved. This forms a positive feedback loop of data and model performance. In this way, we obtain a large-scale pre-trained model for a specific domain. This model can support multiple downstream intelligent tasks such as visual question answering, image captioning, image classification, image generation, etc.

\begin{figure*}[t]
	\centering
	\vspace{-0.15in}
	\begin{minipage}{1\linewidth}	
		\subfigure[Adapter Finetuning]{
			\label{fig:1}
			\includegraphics[width=0.49\linewidth,height=1.5in]{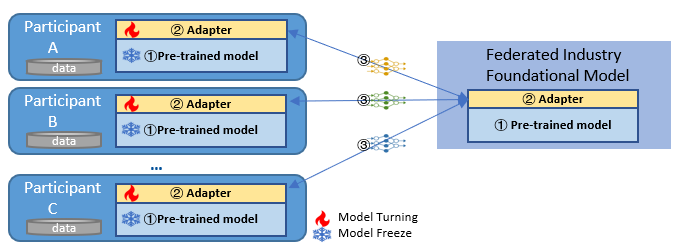}	
		}\noindent
		\subfigure[ Feature Map Knowledge Distillation]{
			\label{fig:2}
			\includegraphics[width=0.49\linewidth,height=1.5in]{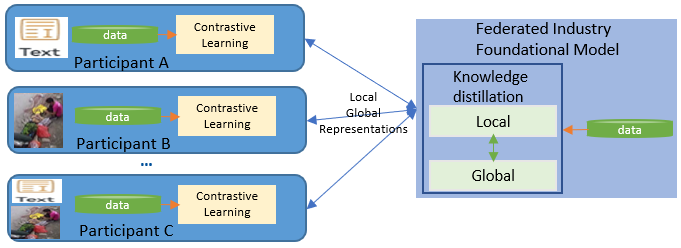}
		}
	\end{minipage}
	\vskip -0.3cm 
	\begin{minipage}{1\linewidth }
		\subfigure[Representation Space Allignment]{
			\label{fig:3}
			\includegraphics[width=0.49\linewidth,height=1.2in]{{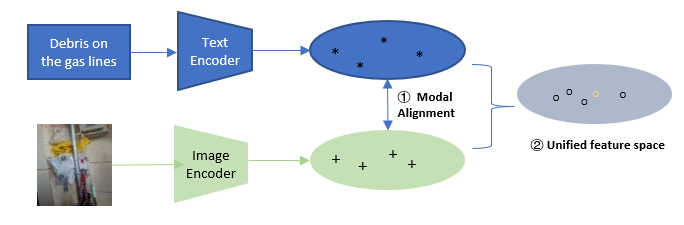}
			}
		}\noindent
		\subfigure[Network Bridging and Connection]{
			\label{fig:4}
			\includegraphics[width=0.49\linewidth,height=1.2in]{{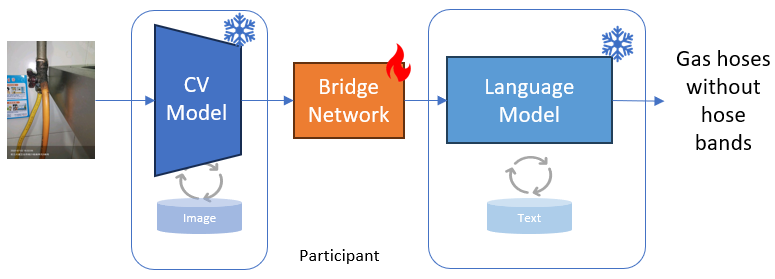}
			}
		}
	\end{minipage}
	\vspace{-0.1in}	
	\caption{The types of aggregation for the federated multimodal model.}
	\vspace{-0.2in}		
	\label{fig:1234}
\end{figure*}

\subsection{Heterogeneous Model Fusion}
Multimodal large-scale models trained on massive public datasets are becoming increasingly large, while the participants in federated learning have relatively limited and unimodal private data, which can lead to model forgetting and overfitting, as well as high resource and time costs, when fully fine-tuning the large models. Federated multimodal large-scale models do not need to aggregate all the parameters, and can decide what to aggregate based on the following methods,as shown in Fig \ref{fig:1234}. 
\begin{itemize}
\item $\boldsymbol{Adapter \ Fine\-tuning:}$ By freezing the local open-source general large-scale models and fine-tuning the adapters, the local training computational overhead can be significantly reduced, as well as the amount of model parameters transferred between the participants and the server. For example: LORA\cite{hu2021lora} freezes the parameters of the pre-trained model and adds matrices A and B, and only updates A and B when fine-tuning the downstream tasks.
\item $\boldsymbol{Feature \  Map \  Knowledge \ Distillation:}$. Using federated knowledge distillation, we can handle the heterogeneity of model architectures and data modalities among the participants, while leveraging the single-modal or multi-modal data on the clients to train a larger server-side model. By using additional public datasets as the medium for knowledge distillation, we transfer the feature map of the public data between the server and the clients, which complement the missing modalities for the single-modal clients, constrain the clients to learn a global consensus, and do not expose the private models and data of the clients.
\item $\boldsymbol{Representation \ Space \ Allignment:}$ Ideally, a federated representation space with different types of data can enable the model to learn the information of other modalities. The textual representations learned from large-scale web data can serve as the targets for learning different modalities features, and the federated partners are selected to maximize the diversity of data, which improves the coverage ratio in the label space, and aligns other modalities such as images, sounds, and videos with the textual modality as the “mediator”.
\item $\boldsymbol{Network \ Bridging \ and \ Connection}$. To improve the efficiency and effectiveness of training, the clients can use the pre-trained image encoder and text encoder weights to map images and texts to a better semantic space. Then, an additional bridge network is added, which is responsible for aligning the image and text spaces. Multiple participants jointly train the bridge network to achieve faster modality alignment and represent domain specific knowledge.
\end{itemize}

\subsection{Flexible Model Aggregation}
Transformer-based deep-learning network has been verified to be more robust than those with a traditional CNN architecture, through prior research and numerous experiments\cite{qu2022rethinking} \cite{hendrycks2020augmix}\cite{bhojanapalli2021understanding}. They can alleviate catastrophic forgetting of data and enhance model convergence. In federated learning, networks with a Transformers architecture achieve higher similarity and synchronization efficiency among the participants\cite{fedvit}. Therefore, multimodal networks using Transformer as the backbone can fuse multi-source heterogeneous data more efficiently and prevent catastrophic forgetting. In the era of big models, federated learning does not require the participants to frequently and iteratively update and aggregate the model. Multiple enterprises can adopt asynchronous\cite{xie2019asynchronous}, chained\cite{chang2018distributed}, or continual learning\cite{lesort2020continual} and obtain the expected model performance. This more relaxed collaboration mechanism can substantially reduce the challenges and costs of actual deployment, as well as ease the participants’ concerns about network security and other issues.

\subsection{Data Quality Improvement}
\begin{figure}
\centering
\includegraphics[width=0.45\textwidth]{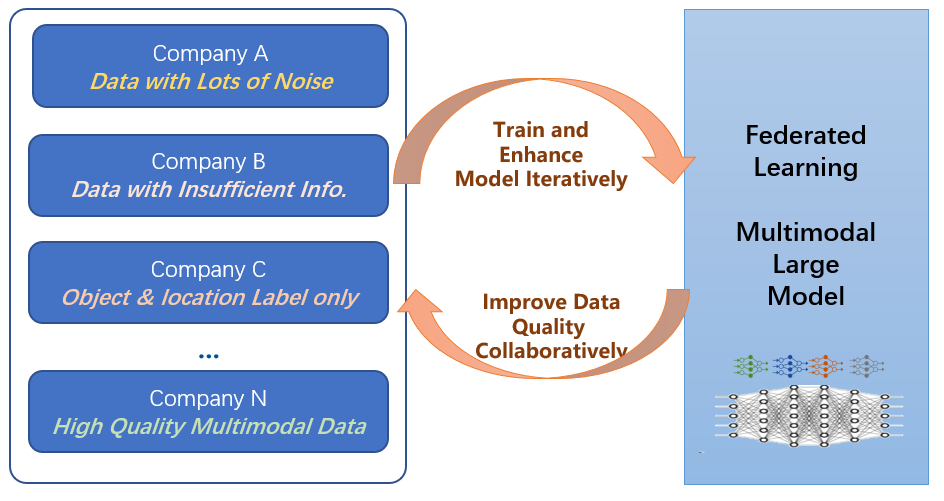}
\caption{Cross-enterprise data cleaning method.}
\label{fig:data_quality}
\end{figure}

One of the crucial factors for developing effective multimodal large models in the industrial internet domain is the availability and quality of multimodal data. Increasing the size of the dataset and including more comprehensive and varied samples from different modalities can enhance the performance of multimodal large models. However, in the industrial internet domain, data from different industries and modalities are distributed among different enterprises, and they cannot be aggregated and processed due to confidentiality and other constraints. To address this challenge, we propose a cross-enterprise data cleaning method that utilizes the power of large models and federated learning technology,as illustrated in Fig \ref{fig:data_quality}.The proposed data cleaning method consists of the following steps:

\begin{enumerate}
  \item[(1)] An initial model is trained by federated learning on the data from N  partners. However, due to the heterogeneous and noisy nature of the training data, the initial model fails to achieve satisfactory performance for practical use.
  \item[(2)] The initial model is then used to screen the data from N participants, and eliminate the low-quality data, retaining only the high-quality data. This enhances the quality of the participant data.
  \item[(3)] The high-quality data obtained in (2) are used to further train a more effective model by federated learning.
  \item[(4)]  Steps (2) and (3) are iterated until the performance of the trained model meets the expected criteria.
\end{enumerate}

\subsection{Data Privacy Preserving}

Based on the strategy of aggregating large models, it's crucial to re-evaluate potential risks related to privacy leakage. It's also essential to implement cost-efficient methods to preserve privacy. Some effective solutions include:
\begin{itemize}
    \item Engaging with Trusted Execution Environments, which offer hardware-based isolation, enabling the refinement of general models using industry-specific private data on a reliable platform.
    \item  Splitting Learning and Knowledge Distillation, which permits part of the model training locally and then sending the feature map (or embedding) to a cloud platform for the main large model training phase.
    \item Leveraging cryptography methods, such as Multi-Party Computation (MPC), Homomorphic Encryption (HE), and Differential Privacy (DP). These involve considerable overheads in communication and computation, as well as introduce data noise for added security.
 \end{itemize}   
 
Additionally, measures must be taken to minimize the risk of unintentionally revealing business-sensitive information via the outputs of large models.

\begin{itemize}
    \item Sanitizing Training Data: It's vital to ensure that the model learns from data free from private or delicate details, like user IDs or precise addresses.
    \item Security Fence and Filtering: A proactive strategy for preventing large model leaking privacy is to filter and monitor their results. This includes blacklisting specific words or phrases, integrating post-processing protocols, and using conditional generation techniques.

\end{itemize} 

\subsection{Incentive Mechanism}

Shapley Value (SV) is a well-known approach to evaluate individual's marginal contribution in a coalition, but the canonical SV calculation and its available variants are very costly.  Our proposed WTDP-Shapley method for FL contribution measurement has been proven to be effective and efficient in a real city-gas household inspection scenario \cite{WTDP-Shapley}. Using WTDP-Shapley allows for a quick assessment of each participant's contribution with a guaranteed low error rate.However, in big model era, there are several challenges in applying the WTDP-Shapley method for measuring sharing.
\begin{itemize}
    \item Although the performance has been improved dramatically, the cost is still un-affordable to generate and evaluate a group of large models with billions of parameters.
    \item Contribution evaluation should decompose the model structure, and consider the corresponding data modality and model blocks. We can address these challenges by employing a dynamic masking approach, which involves masking different block weights to identify the block that play a significant role in the prediction at each time step. This approach highlights the important block that vary over time, allowing us to assess their contributions\cite{crabbe2021explaining}.
    \item Since the foundation large model has been pre-trained based on public datasets with general knowledge, data uniqueness and domain knowledge would become important to bring added values for domain-specific model. 
\end{itemize}

\section{Case Study: Multimodal Large Model for Urban Operation Management}
\label{CaseStudySection}
Urban operation management requires a holistic solution, based on various data sources (e.g., surveillance video and inspection photo) and knowledge (e.g., reports and regulations) from different companies and government agencies. Therefore, a domain-specific multimodal large model is desired to inspect safety hazardous in any corner of the city regarding to infrastructure and human behaviours. This comprehensive large model helps to improve the urban safety and efficiency. By integrating visual, textual, auditory, and other modality data, the model can perform real-time monitoring and early warning, and thus help the urban management departments to quickly respond and handle various incidents, ensuring the safety of residents’ lives and property. In addition, it is also very helpful for city management and planners to improve facility layout and resource scheduling for energy saving, as well as living convenience and quality.

As urban operation management is a complex and dynamic process, various data sources and domain expertise are required to train a multimodal large model covering the entire application scenarios. To solve this problem, ENN Group, WiNDAKA, and  UNICOM(Shanghai) have collaborated to establish a collaborative ecosystem. As illustrated in Table~\ref{tab:data_summary}, the three companies have expert knowledge and rich experience in energy infrastructure safety, residential community security, and urban management. They have accumulated large amounts of data over the years to support a vision-language large model training for urban operation and management vertical domain. The multi-party collaboration is essential for covering smart city intelligence tasks. The parties share and integrate complementary data and knowledge, forming a data aggregation effect, thereby establishing a more comprehensive and accurate multimodal large model.

\begin{table*}[htbp] 
  \centering
  \caption{Overview of Enterprise Data and Scenarios}
  \begin{tabular}{p{0.15\linewidth}p{0.3\linewidth}p{0.2\linewidth}p{0.15\linewidth}p{0.1\linewidth}}
    \toprule
    Company & Scenario Description & Data Type and Quantity & Annotation Present \\
    \midrule
    ENN Group & Energy Security: Hazard Identification and Descriptive Data & Visual and Textual, 260k+ & Yes \\
    Unicom (Shanghai)  & Urban Management: City Streets and Shop Management & Visual and Textual, 150k+ & Partial \\
    WiNDAKA & Community Security: Attributes of People and Vehicles & Image, 110k+ & Yes \\
    \bottomrule
  \end{tabular}
  \label{tab:data_summary}
\end{table*}

However, challenges should be tackled to establish a multimodal large model successfully in federated learning manner:
\begin{itemize}
\item Instead of training from scratch, we need to choose a suitable pre-trained vision-language general model as the foundation. It is able to fuse multimodal data relevant to smart city, support mainstream visual-text tasks, and have strong Chinese understanding and output capabilities.
\item A efficient and robust federated learning framework should be deployed securely and quickly amongst multiple parties, supporting various algorithms for model aggregation, data security and privacy protection.
\item The quality of multimodal data in each party should be improved. For instance,  the existing image caption is non-informative and incomplete. However, professional expert annotation cost is very high to improve data quality manually.
\item As the model size has been increase significantly, federated learning requires huge communication bandwidth, which is very costly especially when the companies  must be connected via VPN for security and fault-tolerance considerations.
\end{itemize}

To address the four challenges mentioned above, we propose corresponding solutions in following subsections.
\subsection{Unified multimodal network}
Multimodal learning primarily encompasses two directions, as shown in Fig \ref{fig:bothFigures}:

\begin{figure}[htbp]
\centering
\subfigure[Multimodal feature fusion with multi-tower structure.]{\includegraphics[width=0.5\linewidth]{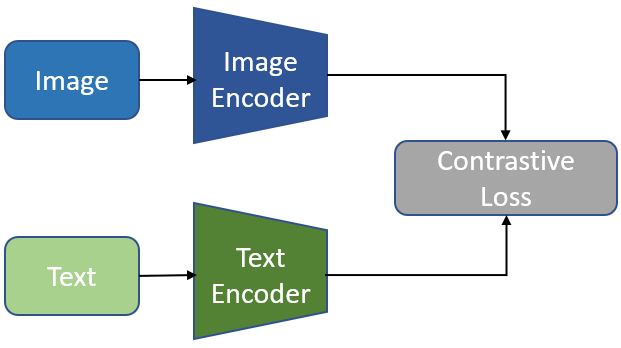}\label{fig:dual}}
\hfil
\subfigure[Multimodal information with large language
model.]{\includegraphics[width=0.4\linewidth]{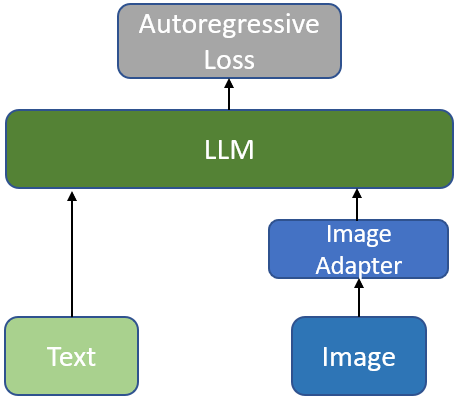}\label{fig:llm}}

\caption{Two directions of multimodal learning.}
\label{fig:bothFigures}
\end{figure}

\begin{itemize}
\item Multimodal feature fusion with multi-tower structure. In the early days of self-supervised learning, there were no powerful language models or visual base models. The work at this stage mainly focused on using various self-supervised learning objectives, designing different feature fusion networks, and performing end-to-end pre-training, in order to obtain multimodal pre-trained models. The main representative works include CLIP\cite{radford2021learning},ALBEF\cite{li2021align}, BLIP\cite{li2022blip},  etc.

\item Incorporating multimodal information into large language models. The main representative works include OFA\cite{wang2022ofa}, Flamingo\cite{alayrac2022flamingo}, Palm-E\cite{driess2023palm}, X-LLM\cite{chen2023xllm}, UnIVAL\cite{shukor2023unified} etc., which usually keep the parameters of LLM and pre-trained visual models freezed, and add extra trainable modules to unify a large number of multimodal tasks into generation tasks, and perform pre-training under a unified self-regressive pre-training paradigm.
\end{itemize}

Alibaba DAMO Academy has introduced a Seq2Seq generative framework OFA\cite{wang2022ofa}, which unifies modality, task, and structure. OFA shares the same model structure across all tasks, and uses artificially designed instructions to distinguish tasks. It can cover downstream tasks spanning multimodal generation, multimodal understanding, image classification, natural language understanding, text generation and other scenarios, and achieves state-of-the-art results on multiple tasks. At the same time, OFA has been trained on a large amount of multimodal Chinese corpus, and has strong Chinese semantic understanding and output capabilities.

\subsection{ENNEW Federated Learning platform}
ENNEW Federated Learning platform, which was certified by China Academy of Information and Communications Technology in 2022, can meet the federated learning basic capability  requirements in terms of scheduling management, data processing, algorithm implementation, effect and performance, security, etc. It adopts TLS encryption technology to ensure data is always protected during transmission and storage, strengthens identity verification and permission management mechanisms to prevent malicious parties from accessing and tampering with other parties’ data, and introduces multi-party secure computing technology to support multi-party model secure aggregation and prevent model parameters and gradient information from reverse deducing confidential information. In addition, it is flexible in terms of distributed deployment and network policy. For example, the model parameter aggregation server provides RESTFul API and message queue services, and each party does not need to open separate services and network ports. 
\begin{figure*}[t]
\centering
\includegraphics[width=1\textwidth]{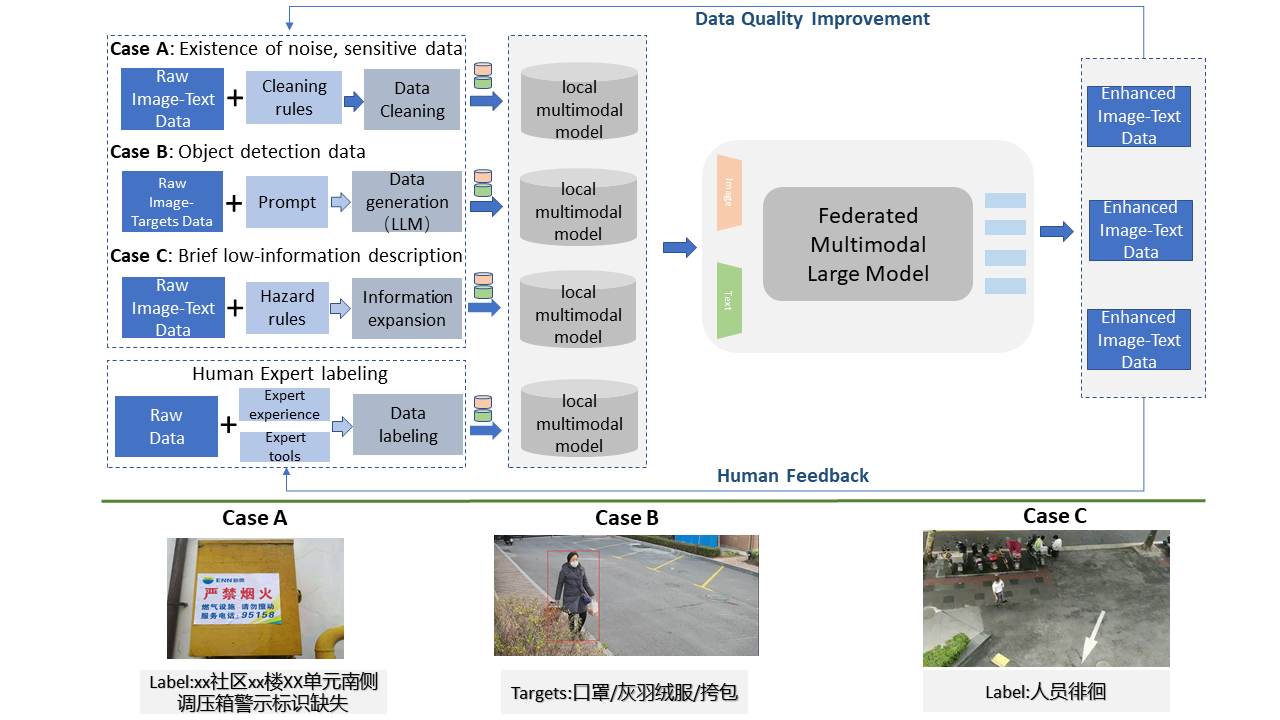}
\caption{Approaches for improving data quality in different cases.}
\label{fig:data_enhance}
\end{figure*}

\subsection{Improving data quality}
In the industrial internet, due to historical management and other reasons, the sample data generation has a lot of randomness and incompleteness, which becomes the bottleneck for each enterprise to achieve industrial digitalization. However, it is very costly in terms of human and time resources to re-annotate all the data with domain experts,as shown in Figure \ref{fig:data_enhance}. We adopt a model self-iteration data augmentation scheme, where data and model mutually promote each other, and generate more realistic and comprehensive image captions.For Case A where there is a lot of noise such as address information and sensitive information in the data, we need to perform data cleaning based on the rules. For data with only images and corresponding object labels, we need to use LLM to assist in generating some descriptive text by using prompts.For data with text-image pairs, but the text descriptions are too short to cover the complete scene, we need to add some custom rules to perform text expansion.  In different scenarios, multimodal models are first trained locally, and a federated multimodal large model is formed through federated learning, which aggregates data and knowledge from multiple parties. The original local data can be filtered and augmented using a filter-based approach to improve data quality. At the same time, high-quality data can be generated through expert annotation feedback. Through continuous iteration, both the data and model performance can be enhanced.
\subsection{Communication efficiency}
 In the era of large models, the scale of pre-trained large models becomes more larger, and federated learning with full transmission requires a lot of communication bandwidth, making communication a bottleneck and an impossible solution. At the same time, it is also more difficult to perform full fine-tuning. The parameter-efficient fine-tuning method PEFT fixes most of the pre-trained parameters and only fine-tunes a small number or additional model parameters, greatly reducing the computation and storage costs. The PEFT scheme represented by LORA only fine-tunes a small number of parameters in the model, reducing the parameter training overhead to one percent of the original, and achieving very considerable results. We adopt the parameter-efficient fine-tuning technology PEFT, and only upload the fine-tuned parameter part in the federated learning process.
 \subsection{performance evaluation standards}
Academic performance evaluation standards for large models, such as BLEU, METEOR, ROUGE, CIDEr, etc., measure the performance of models by comparing the difference between the generated image caption and the ground truth. However, federated learning not only enhances the original intelligence capabilities of the parties, but also has the intelligence capabilities of other parties, and through powerful inference capabilities, superimposes and integrates the intelligence capabilities of each party to achieve a higher level of intelligence. Obviously, it is unreasonable to compare the results generated by federated learning with the ground truth of a party’s limited knowledge. Therefore, it is necessary to form a validation set of large models in vertical domains and re-annotate them in a joint intelligence manner. Since this evaluation method requires re-annotation of the existing validation set, it requires manpower to annotate, which is time-consuming and laborious. Currently, subjective evaluation methods or GPT4 are used to score, and a certain number of validation images are selected to evaluate whether the multimodal large model generated by federated learning has produced joint intelligence.





\subsection{Experiment Results}

To be updated with the latest experimental results, including the adoption of LORA in federated learning framework for cost-saving multimodal large model fine-tuning, the performance evaluation of the federated leaning vision-language model in terms of benchmark metrics (e.g., Rouge) and its advantages on generating professional and informative image captions in various application scenarios.

\section {Conclusion}
\label{ConclusionSection}
The evolution of large AI models like ChatGPT highlights the importance of multimodal data for advancing general AI. However, most existing general multimodal large models were trained public datasets and their performance in specific industrial domains are under expectations. In this paper, we have proposed a federated learning framework in big model era, supporting companies to fine-tune a general multimodal model using their proprietary data and knowledge in a collaborative and efficient manner. The framework has been deployed in companies from energy infrastructure, residential community property management and city operation management. The preliminary experimental results has shown that the vision-language large model could learn from diverse and heterogeneous data from companies and integrate their expert knowledge for understanding images better from multiple perspectives. This is essential for establishing a holistic solution to urban operation and management, inspiring novel business model for safety, energy saving, environment-friendly high-quality city living. As the federated learning ecosystem grows, it promises to accelerate collaborative intelligence for industry domain-specific large models and bring researchers together to tackle the challenges of federated learning in big model era, in terms of novel approaches to fuse heterogeneous data and model, ensuring data privacy and security based on the understanding of large models, enhance training efficiency using creative fine-tuning and model aggregation methods, incentive mechanism and IP protection methods for sustainable ecosystem collaboration, and etc.
\bibliographystyle{ieeetr} 
\bibliography{reference}
\

\end{document}